\documentclass[runningheads]{llncs}
\usepackage{graphicx}
\usepackage{subcaption}
\usepackage{boldline}

\usepackage{caption}
\usepackage{xcolor,colortbl}
\usepackage{soul}
\usepackage{pgfplots}
\pgfplotsset{compat=1.12}
\usepackage{multirow}
\usepackage{todonotes}
\usepackage[hyphens]{url}
\usepackage{hyperref}
\begin{document}
%
\title{When Machine Learning Models Leak: An Exploration of Synthetic Training Data}

\titlerunning{When Machine Learning Models Leak}
%

\author{Manel Slokom\inst{1,2,3} \and Peter-Paul de Wolf \thanks{The views expressed in this paper are those of the authors and do not necessarily reflect the policy of Statistics Netherlands.} \inst{2} \and Martha Larson \inst{3} }
\institute{Delft University of Technology, The Netherlands \\ \email{m.slokom@tudelft.nl} \and Statistics Netherlands, The Hague, The Netherlands \\ \email{pp.dewolf@cbs.nl}\and Radboud University, The Netherlands \\ \email{m.larson@cs.ru.nl}}

\authorrunning{M. Slokom et al.}

\maketitle              
\begin{abstract}
We investigate an attack on a machine learning model that predicts whether a person or household will relocate in the next two years, i.e., a propensity-to-move classifier. 
The attack assumes that the attacker can query the model to obtain predictions and that the marginal distributions of the data set on which the model was trained are publicly available. 
The attack also assumes that the attacker has obtained the values of non-sensitive attributes for a certain number of target individuals. 
The objective of the attack is to infer the values of sensitive attributes for these target individuals. 
We explore how replacing the original data with synthetic data when training the model impacts how successfully the attacker can infer sensitive attributes.\footnote{This paper is a corrected and updated version of the original paper, which was published as: Slokom, M., de Wolf, PP., Larson, M. (2022). When Machine Learning Models Leak: An Exploration of Synthetic Training Data. In: Domingo-Ferrer, J., Laurent, M. (eds) Privacy in Statistical Databases. PSD 2022. Lecture Notes in Computer Science, vol 13463. Springer, Cham.}

\end{abstract}
\section{Introduction}
\label{sec:intro}
Governmental institutions charged with collecting and disseminating information may use machine learning (ML) models to produce estimates, such as imputing missing values or inferring attributes that cannot be directly observed. 
When such estimates are published, it is also useful to make the machine learning model itself publicly available, so that researchers using the estimates can evaluate it closely, or even produce their own estimates.
Moreover, society also asks for more insight into the models that are used, e.g., to address possible discrimination caused by decisions based on machine learning models. 

Unfortunately, machine learning models can be attacked in a way that allows an attacker to recover information about the data set that they were trained on~\cite{Liu2021When}.
For this reason, making machine learning models available can lead to a risk that information from the training set is leaked.
In this paper, we carry out a case study of \emph{model inversion attribute inference attacks} on a machine learning classifier to better understand the nature of the risk. 
Model inversion attribute inference attacks aim to reconstruct the data a model is trained on or expose sensitive information inherent in the data~\cite{hidano2020exposing,wang2021variational}.
Conventionally, they only seek to infer sensitive attributes of individuals whose data are included in the training set (Inclusive individuals). 
Here, we go beyond this conventional perspective to investigate the extent to which the availability of the machine learning model and the marginal distributions of the data it was trained on can support inferring sensitive attributes of individuals who are not in the training set (Exclusive individuals).

The attack scenario that we study assumes that the classifier has been made accessible and can be queried with arbitrary input an unlimited number of times, and also that the marginal distributions of the data set the model was trained on have been released. 
The attacker has a set of non-sensitive attributes of the target individuals including the correct value for the propensity to move attribute for ``Inclusive individuals'' in the training data or ``Exclusive individuals'' not in the training data.
The attacker wishes to learn sensitive attributes for a group of victims, i.e., target individuals.
The classifier that we attack in our study predicts individuals' tendency to move or relocate, i.e., whether an individual or household has the desires, expectations, or plans to move to another dwelling~\cite{crull1979residential} within the next two years. 
For this reason, it is called a \emph{propensity-to-move} classifier.
Our investigation into propensity to move builds upon the work of~\cite{Burger2019Replacing}, which studies the possibility of replacing a survey question about moving desires with a model-based prediction using a machine learning classifier.

Our experimental investigation first confirms that a machine learning classifier is able to predict the propensity to move with an accuracy comparable to that achieved by~\cite{Burger2019Replacing}.
In contrast to~\cite{Burger2019Replacing}, we report results for previously ``unseen'' individuals (Exclusive individuals) separately from results on individuals who appeared in the training data, which was collected two years before the test data (Inclusive individuals).
We then attack this classifier and demonstrate that an attacker can learn sensitive attributes both for Inclusive individuals in the training data as well as for Exclusive individuals.
Next, 
we train the machine learning classifier on synthetic training data and repeat the attacks.
The resulting classifier is slightly less susceptible to attacks compared to the original classifier, which was trained on the original data.
Our findings point in the direction that future research must pursue to investigate other model inversion attribute inference attacks, as well as other synthetic data techniques that could further reduce the risk of attacks when used to train machine learning models.

\section{Threat Model}
\label{sec:threat}
Our goal is to be able to make publicly accessible a machine learning model that has been trained on synthetic data such that the model maintains the same performance as a model trained on the original data, but is less susceptible to model inversion attribute inference attacks.
In this section, we specify our goal more formally in the form of a threat model.

Inspired by~\cite{Salter:1998}, we include four main dimensions in our threat model. 
First, the threat model describes the adversary by looking at the resources at the adversary's disposal and the adversary's objective.
In other words, it specifies what the attacker is capable of and what the attacker's goal is. 
Second, it describes the vulnerability, including the opportunity that makes an attack possible.
Then, the threat model specifies the nature of the countermeasures that can be taken to prevent the attack.

Table~\ref{tab:threatCBS} provides the specifications of our threat model for each of the dimensions.
As objective, the attacker seeks to infer specific sensitive attributes of the target individuals.
As resources, we assume that the attacker has collected a set of non-sensitive attributes of the target individuals, i.e., previously released data or data gathered from social media.
The target individuals are either in the training data used to train the released model (``Inclusive individuals'') or not in the training data (``Exclusive individuals'').
The set of non-sensitive attributes also includes the target individuals' corresponding true labels concerning their propensity-to-move.
The vulnerability is related to the opportunities available to the attacker, i.e., how the model is released and the access that has been provided to the model.
The attacker is able to query the model and collect the output predictions of the model, for an unlimited number of arbitrary inputs.
The attacker also has information about the marginal distribution for each attribute in the training data.
Finally, the countermeasure that we are investigating is modification of the training data on which the model is trained.

\begin{table}[]
\centering
\caption{Threat model addressed by our approach.}
\label{tab:threatCBS}
\resizebox{\columnwidth}{!}{%
\begin{tabular}{ll}
\hline
\multicolumn{1}{c}{\textbf{Component}} & \multicolumn{1}{c}{\textbf{Description}} \\ \hline
\textit{Adversary: Objective} & Specific sensitive attributes of the target individuals. \\ \hline
\textit{Adversary: Resources} & \begin{tabular}[c]{@{}l@{}}A set of non-sensitive attributes of the target individuals, \\ 
including the correct value for the propensity-to-move \\ attribute,
for ``Inclusive individuals'' (in the training set) \\ 
or ``Exclusive individuals'' (not in the training set).\end{tabular} \\ \hline
\textit{Vulnerability:Opportunity} & \begin{tabular}[c]{@{}l@{}}Ability to query the model to obtain output plus\\ the marginal distributions of the data set that the \\ model was trained on. \end{tabular} \\ \hline
\textit{Countermeasure} & Modify the data on which the model is trained. \\ \hline
\end{tabular}%
}
\end{table}

\section{Background and Related Work}
\label{sec:background}
In this section, we give a brief overview of basic concepts and related work on predicting propensity to move, privacy in machine learning, and model inversion attribute inference attacks.

\subsection{Propensity to Move}
``Propensity to move'' is defined as desires, expectations, or plans to move to another dwelling~\cite{crull1979residential}.
Multiple factors come into play to understand and estimate the propensity to move in a population. 
In~\cite{crull1979residential}, the authors have grouped those factors into two categories: 
(1) \textit{Residential satisfaction}, which is defined as the satisfaction with the dwelling and its location or surroundings.
Residential satisfaction is divided into housing satisfaction and neighborhood satisfaction.
(2) \textit{Household characteristics}, which is related to demographic and socioeconomic characteristics of the household.
Gender and age are indicators of a household are important demographic attributes.
For instance, a male household has different mobility patterns than a female household. 
Also, education and income of the household are important socioeconomic attributes.

In~\cite{kleinhans2009does}, the authors studied the social capital and propensity to move of four different resident categories in two Dutch restructured neighborhoods. 
They define social capital as the benefit of cursory interactions, trust, shared norms, and collective action. 
Using a logistic regression model, they show that (1) age, length of residency, employment, income, dwelling satisfaction, dwelling type, and perceived neighborhood quality significantly predict residents' propensity to move and
(2) social capital is of less importance than suggested by previous research.
In~\cite{fackler2017losing}, the authors investigate the possible relationship between involuntary job loss and regional mobility.
The authors look at whether job loss increases the probability of relocating to a different region and whether displaced workers who relocate to another region after job loss have better labor market outcomes than those staying in the same area.
They find that job loss has a strong positive effect on the propensity to relocate.
In~\cite{coulter2015motivates}, the authors use data collected by the British Household Panel Survey.
The authors tested seven hypotheses to examine the reasons why people desire to move and how these desires affect their moving behavior.
The results show that people are more likely to relocate if they desire to move for targeted reasons like job opportunities than if they desire to move for more diffuse reasons relating to area characteristics.
In~\cite{Burger2019Replacing}, the authors study the possibility of replacing a survey question about moving desires with a model-based prediction.
To do so, they use machine learning algorithms to predict moving behavior.
The results show that the models are able to predict the moving behavior about equally well as the respondents of the survey.
In~\cite{de2020later}, the authors examine the residential moving behavior of older adults in the Netherlands.
The authors of~\cite{de2020later} use data collected from Housing Research Netherlands (HRN) to provide insights into the housing situation of the Dutch population and their living needs.
A logistic regression model was used to assess the likelihood that respondents would report that they are willing to move in the upcoming years.
They show that older adults are more often motivated by unsatisfactory conditions in the current neighborhood.
Here, we follow up on the work of~\cite{Burger2019Replacing}, as they evaluate a number of machine learning models to predict the propensity-to-move.

\subsection{Privacy in Machine Learning}
In this section, we will discuss challenges and possible solutions for privacy in machine learning.
Existing works can be divided into three categories according to the roles of machine learning (ML) in privacy~\cite{Liu2021When}:
First, \textit{making the ML model private}.
This category includes keeping the ML model (its parameters) confidential and protecting the data it was trained on to control privacy threats.
Second, \textit{using ML to enhance privacy protection}.
In this category, ML is used as a tool to enhance the privacy protection of the data.
Third, \textit{ML based privacy attack}.
The ML model is used as an attack tool by the attacker.

Our work falls under the first category.
The governmental institution that wishes to make the model available to the public needs to protect individuals in the training data and to make sure that by providing access to the model they are not making it easier to infer sensitive information about individuals not in the training data.
As mentioned in Section~\ref{sec:intro}, we approach the protection of the model by leveraging synthetic data. 
In prior work~\cite{Abadi2016DLDP}, the authors propose a one-step approach for differential private (DP) training of neural networks.
They introduce differential private stochastic gradient descent (DPSGD) that achieves differential privacy by constraining the gradients to have a maximum $l_{2}$ norm for each layer. 
Existing DP research mainly focuses on protecting against membership inference attacks.
In contrast, we employ a two-step approach in which we first synthesize data and then proceed to train a machine learning model.

\subsection{Synthetic data generation} 
\label{sub:synthetic}
Synthetic data generation is based on two main steps: First, we train a model to learn the joint probability distribution in the original data.
Second, we generate a new artificial data set from the same learned distribution. 
In recent years, advances in machine learning and deep learning models have offered us the possibility to learn a wide range of data types.

Synthetic data was first proposed for Statistical Disclosure Control (SDC)~\cite{drechsler2011synthetic}.
The SDC literature distinguishes between two types of synthetic data~\cite{drechsler2011synthetic}.
First, \textit{fully synthetic data sets} create entirely synthetic data based on the original data set.
Second, \textit{partially synthetic data sets} contain a mix of original and synthetic values. 
It replaces only observed values for attributes that bear a high risk of disclosure with synthetic values.

In this paper, we are interested in fully synthetic data.
For data synthesis, we used an open source and widely used R toolkit: \textit{Synthpop}. 
We use a CART model for synthesis since it has been shown to perform well for other types of data~\cite{drechsler2011empirical}.
Data synthesis is based on sequential modeling by decomposing a multidimensional joint distribution into conditional and univariate distributions. 
The synthesis procedure models and generates one attribute at a time, conditionally to previous attributes:
\begin{equation}
f_{x_{1}, x_{2}, ..,x_{n}} = f_{x_{1}} \times f_{x_{2}| x_{1}} \times .. \times f_{x_{n}| x_{1},x_{2},..x_{n-1}} 
\end{equation}
Synthesis using the CART model has two important parameters.
First, the order in which attributes are synthesized is called the \textit{visiting sequence}. 
This parameter has an important impact on the quality of the synthetic data since it specifies the order in which the conditional synthesis will be applied.
Second, the \textit{stopping rules} that dictate the number of observations that are assigned to a node in the tree.

\subsection{Model Inversion Attribute Inference Attack}
\label{sub:InfAttack}
Model inversion attribute inference attacks try to recover sensitive features or the full data sample based on output labels and partial knowledge of some features~\cite{rigaki2020survey,mehnaz2022your}.
In~\cite{Fredrikson2014Privacy,Fredrikson2015Model}, the authors introduce two types of model inversion attacks: black-box attacks and white-box attacks.
The difference between a black-box attack and a white-box attack lies in the resources that are available to the adversary.
In a black-box setting, the adversary can only query the model and receive predictions~\cite{Fredrikson2014Privacy}.
The authors of \cite{Fredrikson2014Privacy} show that an attacker can use a trained classifier to extract representations of the training data.
They exploit access to a model to learn information about its training data using confidence scores revealed in predictions.
In~\cite{mehnaz2022your}, the authors provide a summary of possible assumptions about the adversary's capabilities and resources for different model inversion attribute inference attacks.
The authors propose two types of model inversion attacks:
(1) confidence score-based model inversion attack (CSMIA) and (2) label-only model inversion attack (LOMIA). 
The first attack, CSMIA, assumes that the adversary has access to the target model's confidence scores.
The second attack, LOMIA, which is the basis of our work, assumes that the adversary has access to the target model's predictions only.
The LOMIA attack uses an auxiliary machine learning model to infer sensitive information about target individuals. 
In our attack, we employ LOMIA to access the model's predictions, but we do not use an auxiliary model. 
We opt for LOMIA because it assumes the same adversary resources as in our threat model (Section~\ref{sec:threat}).
Other attacks such as~\cite{Hidano2017Model} assume that the attacker does not have access to target individuals' non-sensitive attributes. 

\subsection{Attribute Disclosure Risk}
In the context of statistical disclosure control, model inversion attribute inference attacks pose a risk to attribute disclosure. 
An adversary leverages predictive models (target ML model) to infer sensitive information about individuals from known attributes, increasing the likelihood of disclosure.
Prior research on attribute disclosure risk has looked at various metrics to measure the risk of attribute disclosure. 
These metrics include matching probability, where perceived match risk, expected match risk, and true match risk are compared~\cite{Reiter2009Estimating}. 
Additionally, a Bayesian estimation approach has been considered, where an attacker is assumed to seek a Bayesian posterior distribution~\cite{Reiter2014Bayesian}. 
Correct Attribution Probability (CAP) is another metric used to measure the risk of disclosure. 
CAP measures the proportion of matches between records from the original data and records from the protected data. 
Here, the protected data refers to the data that the adversary has used to query the target model accompanied by the inferred information. 
CAP calculates the ratio of correct attributions to total matches for a given individual~\cite{Mark2014CAP,Taub2018Differential}.
Methods for measurements of success are discussed in~\cite{Andreou2017IdentityvsAttribute}.

In the context of machine learning, we study attribute disclosure or attribute inference attacks as predictions.
An attacker trains an auxiliary model to predict the value of an unknown sensitive attribute from a set of known attributes given access to raw or synthetic data~\cite{stadler2020synthetic,Hittmeir2020Baseline}.
In this paper, we evaluate the success of our attack following~\cite{mehnaz2022your}, which measures the difference between the adversary's predictive accuracy given the model and the accuracy that could be achieved without the model.
We consider our attack successful when its predictive accuracy surpasses that of a baseline using Marginals Only. 
This implies that using more information than just marginal data can reveal sensitive information about target individuals.

\section{Label-Only MIA with Marginals}
In this section, we describe our label-only MIA + Marginals attack (for short LOMIA + Marginals).
LOMIA + Marginals is based on the LOMIA attack proposed by~\cite{mehnaz2022your}.
The attacker aims to predict the value of an unknown sensitive attribute from a set of known attributes. 
To perform the attack, the attacker needs access to the released ML model's predictions, the released marginal distribution representing possible values and actual probabilities for the sensitive attributes in the training data, and a subset of data containing information about target individuals' non-sensitive attributes.
The attacker queries the target model multiple times by replacing the missing sensitive attribute with all possible values. 
To determine the value of the sensitive attribute, we follow Case (1) proposed in~\cite{mehnaz2022your}.
Case (1) states that when the target model's prediction is correct only for a single sensitive attribute value, e.g., $y= y^{'}_{0} \wedge y != y^{'}_{0}$ or $y != y^{'}_{0} \wedge y = y^{'}_{0}$, the attacker selects the sensitive attribute to be the one for which the prediction is correct. 
For instance, when the sensitive attribute is binary, i.e., $K=2$, the attacker will query the model by setting the sensitive attribute value to both \textit{yes} and \textit{no} and leave other attributes unchanged.
When the sensitive attribute is set to \textit{no}, the returned model prediction is $y^{'}_{0}$.
Similarly, when the sensitive attribute is set to \textit{yes}, the returned model prediction is $y^{'}_{1}$.
If $y= y^{'}_{1} \wedge y != y^{'}_{0}$, the attacker predicts \textit{yes} for the sensitive attribute.
Differently, from~\cite{mehnaz2022your}, when the attacker cannot infer the sensitive attribute (for cases where the model's predictions vary across multiple sensitive attribute values, and cases where the model outputs incorrect predictions for all possible sensitive attribute values), we do not use an auxiliary machine learning model.
Instead, the attacker relies on the released marginal distribution to predict the most probable value of the sensitive attribute.

In addition to the LOMIA + Marginals attack model, we also study an attack that uses \textit{Marginals Only}, as a baseline for comparison.
The Marginals Only attack uses the marginals to predict the most probable value of the sensitive attribute.

\section{Experimental Setup}
\label{sec:experiments}
In this section, we describe our data sets, utility measures calculated by applying different machine learning classifiers, and adversary resources.

\subsection{Data Set}
For our experiments, we used existing data collected by~\cite{Burger2019Replacing} related to the propensity to move of individuals in the Netherlands.
The authors of~\cite{Burger2019Replacing} linked various records from the Dutch System of Social Statistical Datasets (SSD). 
The data set has around 150K individuals including 100K individuals drawn randomly from SSD and ~50K individuals are sampled from the Housing Survey 2015 (HS2015) respondents. 
The resulting data set has 700 attributes for each individual: 
(1) ``y01'' the binary target attribute indicating whether (=1) or not (=0) a person moved in year $j$ where j= 2013, 2015. 
The target attribute ``y01'' is imbalanced and dominated by class 0.
(2) time-independent personal attributes,
(3) time-dependent personal, household, and housing attributes,
(4) information about regional attributes.

\subsubsection{Feature Selection}
\label{sub:FeatSelection}
Different from~\cite{Burger2019Replacing}, we applied feature selection to reduce the number of attributes.
Some attributes can be noise and potentially reduce the performance of the models.
Also, reducing the number of attributes helps to reduce the complexity of synthesis and to better understand the output of the ML model.
To do so, we applied \textit{SelectKBest} from \textit{Sklearn}.
We use the $\chi^2$ method as a scoring function.
We selected the top $K=30$ attributes with the highest scores. 
Our final data set contains the 30 best attributes for a total of 150K individuals. 
In addition to the 30 attributes which include age, we added gender (binary) and income (categorical with five categories) as sensitive attributes.
These sensitive attributes will be used in our model inversion attribute inference attack later (Section~\ref{sub:inference}).

\subsubsection{Data Splits} 
As previously mentioned, our propensity to move data was collected in 2013 and 2015.
Following~\cite{Burger2019Replacing}, we use the 2013 data to train the propensity-to-move classifier and the 2015 data for testing.
We call the 2013 data as ``Inclusive individuals (2013)''.
Recall, that the 2015 data contains both individuals who were present in the 2013 data set (''Inclusive individuals (2015)'') and also new ``unseen'' individuals (``Exclusive individuals (2015)'').
We carry out tests on both sets individually.

In the synthesis process, we are interested in protecting the training data of the target propensity-to-move model.
We use the Inclusive individuals 2013 data to train our synthesis model (cf. Section~\ref{sub:synthetic}.
The generated Inclusive individuals (2013) synthetic data is then used as input for training the target propensity-to-move model.

\subsection{Utility Measures}
In this section, we provide a description of the machine learning classifiers used in our experiments, as well as the metrics to evaluate the performance of these classifiers.

\subsubsection{Machine Learning Classifiers}
We selected a number of machine learning algorithms to predict the propensity to move.
The chosen machine learning techniques provide insight into the importance of the attributes and are easy to interpret and understand~\cite{Burger2019Replacing}.

In our experiments in Section~\ref{sub:MLResults}, we used the following classifiers.
\textit{Decision Tree} creates/learns a tree by splitting the training data into subsets based on an attribute value test. 
This process is repeated on each derived subset in a recursive manner.
In \textit{Random Forest}, each tree in the ensemble is built from a sample drawn with replacement (i.e., a bootstrap sample) from the training set.
\textit{Extra Trees} fits a number of randomized decision trees on various sub-samples of the data set and uses averaging to improve the predictive accuracy and control overfitting.
Extra Trees and Random Forest are ensemble methods.
\textit{Na\"{i}ve Bayes} is a probabilistic machine learning algorithm based on applying Bayes' theorem with strong (na\"{i}ve) independence assumptions between the attributes.
\textit{K-nearest neighbors} (KNN) is a non-parametric machine learning algorithm. 
KNN uses proximity to make predictions about the grouping of an individual data point.
We compare the performance of machine learning algorithms to the performance of a Majority-Class classifier using the most frequent strategy as a na{\"i}ve baseline.

\subsubsection{Metrics for Evaluating Performance of ML Classifiers}
Similar to~\cite{Burger2019Replacing} and since our target propensity-to-move attribute is imbalanced, we used:
F1-score, as a harmonic mean of precision and recall score.
Matthews Correlation Coefficient (MCC), and Area Under the Curve (AUC) measure the ability of a classifier to distinguish between categories.

\subsection{Adversary Resources}
\label{sub:AdvResources}
In this section, we describe different resources that are available for the attacker.
As adversary resources, we assume that the attacker has access to a set of non-sensitive attributes of the target individuals (see our threat model in Section~\ref{sec:threat}).
We consider three cases corresponding to three different sets of individuals:

\begin{itemize}
    \item \texttt{Inclusive individuals (2013)}: The attacker has access to data from the year 2013, which aligns with the data used to train the target model.

    \item \texttt{Inclusive individuals (2015)}: The attacker possesses more recent data from 2015, but it corresponds to the same set of individuals used in training the target model. The more recent nature of the data implies that certain (time-sensitive) attributes for specific individuals may have some changes.

    \item \texttt{Exclusive individuals (2015)}: The attacker's data is from 2015, but it pertains to a distinct group of individuals who were not part of the training set for the target model.

\end{itemize}
We created three different data sets for the three cases. 
Exclusive Individuals (2015) includes all available individuals (2904 individuals). 
For the Inclusive individuals (2013), used to train the target ML model as well as to create the synthetic data, and the Inclusive individuals (2015), we have randomly sampled to create data sets of the same size, each containing 2904 individuals.
The attacker has access to the correct value of the propensity-to-move attribute for the target individuals but does not have information about the sensitive attributes of gender, age, and income, which are the objective of the attacker.

Understanding the vulnerability of a model to model inversion attribute inference attacks requires using the right metric to evaluate different attack models.
Since our sensitive target attributes (gender, age, income) are balanced, we used precision, recall, and F1 to measure the effectiveness of the attacks.
Precision measures the ability of the classifier not to label as positive a sample that is negative.
Precision is the ratio of $tp / (tp + fp)$ where $tp$ is the number of true positives and $fp$ is the number of false positives.
Recall measures the ability of the classifier to label positive samples positive
Recall is the ratio of $tp / (tp + fn)$ where $tp$ is the number of true positives and $fn$ is the number of false negatives. 

\section{Experimental Results}
\label{sec:results}
In this section, we turn to discuss our experimental results. 
We start by describing our results of the performance of the machine learning classifier.
Then, we present the results of model inversion attribute inference attacks.

\subsection{Performance of Machine Learning Classifiers}
\label{sub:MLResults}

Table~\ref{tab:ML_psd_aLL} shows our results of classification performance of propensity to move. 
In Table~\ref{tab:ML_psd_aLL}, the first column reports results on a test set that combines Inclusive Individuals (2015) and Exclusive Individuals (2015). 
This test setting is similar to~\cite{Burger2019Replacing}. 
The second column reports results in the case where the test set is Exclusive individuals (2015).
In the latter case, we evaluate the performance of the trained propensity-to-move on unseen individuals who were not part of the training set.

As expected, all classifiers outperform the Majority-Class baseline, with classifiers using trees generally being the stronger performers.
We also see that when the test set includes Inclusive individuals (2015) and Exclusive individuals (2015), the performance is better than when it includes only ``unseen'' individuals (Exclusive individuals (2015)).
Note that if the data for the \emph{inclusive} individuals were identical in the training and test set, we would have expected very high classification scores.
However, the data is not identical because it was collected on two different occasions with two years intervening, and individuals' situations would presumably have changed.

\begin{table}[]
\centering
\caption{Results of the performance of propensity-to-move model trained on \textbf{original data} and \textbf{synthetic data}. The test data is used in its original (unprotected) form.}
\label{tab:ML_psd_aLL}
\resizebox{\columnwidth}{!}{%
\begin{tabular}{cccccccc}
\hline
\multirow{2}{*}{\textbf{Data Sets}} & \multirow{2}{*}{\textbf{\begin{tabular}[c]{@{}c@{}}Machine learning\\ Algorithms\end{tabular}}} & \multicolumn{3}{c}{\textit{\begin{tabular}[c]{@{}c@{}}Test set: Inclusive individuals\\ (2015) and Exclusive \\ individuals (2015)\end{tabular}}} & \multicolumn{3}{c}{\textit{\begin{tabular}[c]{@{}c@{}}Test set: \\ Exclusive individuals (2015)\end{tabular}}} \\ \cline{3-8} 
 &  & \textbf{AUC} & \textbf{MCC} & \textbf{F1-Macro} & \textbf{AUC} & \textbf{MCC} & \textbf{F1-Macro} \\ \hlineB{4}
\multirow{6}{*}{\textbf{\begin{tabular}[c]{@{}c@{}}Original\\ Data\end{tabular}}} & \textit{Majority-Class} & 0.5000 & 0.0012 & 0.4924 & 0.5000 & 0.0000 & 0.3758 \\ \cline{2-8} 
 & \textit{Naive Bayes} & 0.6815 & 0.2204 & 0.5968 & 0.5656 & -0.0368 & 0.3331 \\ \cline{2-8} 
 & \textit{Random Forest} & 0.7532 & 0.2407 & 0.5946 & 0.7881 & 0.3425 & 0.5732 \\ \cline{2-8} 
 & \textit{Decision Tree} & 0.6568 & 0.2292 & 0.5767 & 0.7180 & 0.3478 & 0.6691 \\ \cline{2-8} 
 & \textit{Extra Trees} & 0.7219 & 0.2099 & 0.5764 & 0.7226 & 0.3197 & 0.6325 \\ \cline{2-8} 
 & \textit{KNN} & 0.6717 & 0.1744 & 0.5575 & 0.6723 & 0.2532 & 0.5981 \\ \hlineB{4}
\multirow{6}{*}{\textbf{\begin{tabular}[c]{@{}c@{}}Synthetic\\ Data\end{tabular}}} & \textit{Majority-Class} & 0.5000 & 0.0000 & 0.4900 & 0.5000 & 0.0000 & 0.3758 \\ \cline{2-8} 
 & \textit{Naive Bayes} & 0.6826 & 0.2029 & 0.5734 & 0.5657 & -0.0144 & 0.3629 \\ \cline{2-8} 
 & \textit{Random Forest} & 0.7275 & 0.2426 & 0.5946 & 0.7870 & 0.3432 & 0.5900 \\ \cline{2-8} 
 & \textit{Decision Tree} & 0.6618 & 0.2125 & 0.5762 & 0.7189 & 0.3567 & 0.6728 \\ \cline{2-8} 
 & \textit{Extra Trees} & 0.7177 & 0.2082 & 0.5596 & 0.7233 & 0.3144 & 0.6425 \\ \cline{2-8} 
 & \textit{KNN} & 0.6542 & 0.1637 & 0.5418 & 0.6437 & 0.2027 & 0.5423 \\ \hlineB{4}
\end{tabular}%
}
\end{table}

\subsubsection{Reproducing Burger et al.,'s~\cite{Burger2019Replacing} results}
In Table~\ref{tab:ML_psd_aLL}, results show that all machine learning classifiers outperform the Majority-Class baseline.
Overall we observe that our results are in line with~\cite{Burger2019Replacing} across different metrics.
This confirms that we can still predict individuals' moving behavior at the same level as in~\cite{Burger2019Replacing} even after reducing the number of attributes.

In addition to reproducing~\cite{Burger2019Replacing}, we looked at another prediction model where train and test individuals are exclusive/different.
We found that it is also possible to predict the moving behavior of new individuals from 2015 based on a classifier trained on different individuals from 2013.

\subsubsection{Measuring the utility of synthetic data}
In order to evaluate the quality of synthetic data, we run machine learning algorithms on a synthesized training set (2013 data).
We used $TSTR$ (train on synthetic and test on real)~\cite{heyburn2018machine} evaluation strategy where we train classifiers on 2013 synthetically generated data and we test on 2015 original data. 
Results in Table~\ref{tab:ML_psd_aLL} show that the performance of machine learning classifiers trained on synthetic data is very close and comparable to the performance of machine learning algorithms trained on original data.
This confirms that the synthetic training set can replace the original training set.
In the remainder of the paper, we will focus on the Random Forest model.
We will assume the release of a Random Forest model, as it outperforms other machine learning classifiers.

\subsection{Results of Model Inversion Attribute Inference Attack}
\label{sub:inference}
In this section, we discuss the results of model inversion attribute inference attacks on the propensity-to-move classifiers using gender, age, and income as the sensitive values.
For comparison purposes, we begin by taking a look at the results generated by LOMIA Case (1) without adding the marginals, which we report in Table~\ref{tab:inversionRes}.
Recall that LOMIA Case (1) (cf. Section~\ref{sub:InfAttack}) involves querying the model under attack with versions of the information of the target individual into which all possible values of the sensitive attribute have been substituted.
\#Predicted individuals reports the raw number of individuals for whom this querying process generates a prediction. 
If there is more than one version of the individual's information that produces the same result from the classifier, then that individual cannot be predicted. 
\#Correctly predicted individuals reports the raw number of predictions that are correct.

First, we consider attacks on the ML models trained on original data.
We see that Inclusive individuals (2013) shows the highest count of correct predictions across the conditions
(gender, age, income).
The correct predictions are relatively lower for both Inclusive individuals (2015) and Exclusive individuals (2015).
This observation can be attributed to the use of Inclusive individuals (2013) for training the target ML model.

Then, we consider attacks on the ML models trained on synthetic data.
For the case of Inclusive individuals (2013), we see that the model trained on synthetic data yields substantially fewer predicted individuals and also fewer correctly predicted individuals than the model trained on original data.
For the case of Inclusive individuals (2015) and Exclusive individuals (2015), the prediction scores (considering both LOMIA + Marginals and Marginals Only) are also in general smaller for the model trained on synthetic data than for the model trained on original data.

The comparison in Table~\ref{tab:inversionRes} illustrates that the use of synthetic data to train models is contributing to mitigate leaks, since models trained on synthetic data yield fewer correctly predicted individuals.
However, it is important to note that for all models and all attributes, the vast majority of the 2904 individuals in each test set (i.e., Attacker resources) cannot be predicted with Case (1) and will be predicted using the marginals in the attacks we discuss below.
Also note that while using LOMIA during the attack, the attacker does not have the ground truth necessary to identify correctly predicted individuals and for this reason, the attack proceeds with the predicted individuals given in Table~\ref{tab:inversionRes}.

\begin{table}[!htb]
\centering
\caption{Results of predictions returned from querying the target model using LOMIA Case (1)~\cite{mehnaz2022your}. \#Predicted individuals are the number of predictions returned from querying the target model. \#Correctly predicted individuals represent correctly predicted records among all target individuals.}
\label{tab:inversionRes}
\resizebox{\columnwidth}{!}{%
\begin{tabular}{ccccc}
\hline
\textit{\textbf{\begin{tabular}[c]{@{}c@{}}Attacker \\ resources\end{tabular}}} & \textit{\textbf{\begin{tabular}[c]{@{}c@{}}Target ML\\ trained on\end{tabular}}} & \textit{\textbf{\begin{tabular}[c]{@{}c@{}}Sensitive\\ Attributes\end{tabular}}} & \textbf{\begin{tabular}[c]{@{}c@{}}\# Predicted\\ individuals\end{tabular}} & \textbf{\begin{tabular}[c]{@{}c@{}}\# Correctly \\ predicted individuals\end{tabular}} \\ \hline
\multirow{6}{*}{\textit{\textbf{\begin{tabular}[c]{@{}c@{}}Inclusive \\ individual \\ (2013)\end{tabular}}}} & \multirow{3}{*}{Original} & \textit{Gender} & 92 & 92 \\ \cline{3-5} 
 &  & \textit{Age} & 38 & 37 \\ \cline{3-5} 
 &  & \textit{Income} & 37 & 37 \\ \cline{2-5} 
 & \multirow{3}{*}{\begin{tabular}[c]{@{}c@{}}Synthetic\\ (CART\\ model)\end{tabular}} & \textit{Gender} & 79 & 35 \\ \cline{3-5} 
 &  & \textit{Age} & 17 & 7 \\ \cline{3-5} 
 &  & \textit{Income} & 27 & 6 \\ \hline
\multirow{6}{*}{\textit{\textbf{\begin{tabular}[c]{@{}c@{}}Inclusive \\ individual \\ (2015)\end{tabular}}}} & \multirow{3}{*}{Original} & \textit{Gender} & 86 & 42 \\ \cline{3-5} 
 &  & \textit{Age} & 20 & 7 \\ \cline{3-5} 
 &  & \textit{Income} & 31 & 5 \\ \cline{2-5} 
 & \multirow{3}{*}{\begin{tabular}[c]{@{}c@{}}Synthetic\\ (CART\\ model)\end{tabular}} & \textit{Gender} & 59 & 35 \\ \cline{3-5} 
 &  & \textit{Age} & 28 & 8 \\ \cline{3-5} 
 &  & \textit{Income} & 25 & 4 \\ \hline
\multirow{6}{*}{\textit{\textbf{\begin{tabular}[c]{@{}c@{}}Exclusive\\ individual \\ (2015)\end{tabular}}}} & \multirow{3}{*}{\textit{Original}} & \textit{Gender} & 281 & 148 \\ \cline{3-5} 
 &  & \textit{Age} & 72 & 16 \\ \cline{3-5} 
 &  & \textit{Income} & 58 & 24 \\ \cline{2-5} 
 & \multirow{3}{*}{\begin{tabular}[c]{@{}c@{}}Synthetic\\ (CART\\ model)\end{tabular}} & \textit{Gender} & 124 & 59 \\ \cline{3-5} 
 &  & \textit{Age} & 47 & 13 \\ \cline{3-5} 
 &  & \textit{Income} & 57 & 16 \\ \hline
\end{tabular}%
}
\end{table}
Next, we move to discuss the results of our LOMIA + Marginals attack.
Here, we start with the case of Inclusive individuals (2013). Table~\ref{tab:InferenceResInc2013} summarizes the results of model inversion attribute inference attacks, comparing LOMIA + Marginals and Marginals Only attacks for Inclusive individuals (2013).
Considering attacks on ML models trained on original data, we observe that the LOMIA + Marginals attack out performs the Marginals Only attack.
Considering attacks on ML models trained on synthetic data, we see that the LOMIA + Marginals attack outperforms the Marginals Only attack for the age attribute, whereas it is surpassed by the Marginals Only attack for gender and income attributes.
Recall that we saw in Table~\ref{tab:inversionRes} that the vast majority of the predictions for individuals are carried out with Marginals.
For this reason, we do not expect a large difference between LOMIA + Marginals and Marginals Only and it is not particularly surprising that a Margainals Only attack might sometimes outperform LOMIA + Marginals.
Turning now to comparison, we see in Table~\ref{tab:InferenceResInc2013} that the strongest attack on an ML model trained on original data (in this case LOMIA + Marginals) is always slightly more successful than the strongest attack on an ML model trained on synthetic data (in this case usually Marginals Only).
We emphasize that the difference is very small, but the fact that it is discernible supports the conclusion that training models on synthetic data does have at least a basic potential for fighting ML model leakage.

\begin{table}[]
\centering
\caption{Case of ``Inclusive individuals (2013)'' as adversary resources: Evaluation results for model inversion attribute inference attacks using Marginals Only and LOMIA + Marginals attacks. Standard deviations, indicated by $\pm$, represent variability across ten experiment runs.}
\label{tab:InferenceResInc2013}
\resizebox{\columnwidth}{!}{%
\begin{tabular}{cccccccccccc}
\hline
\multirow{2}{*}{\textit{\textbf{\begin{tabular}[c]{@{}c@{}}Attacker \\ resources\end{tabular}}}} & \multirow{2}{*}{\textit{\textbf{\begin{tabular}[c]{@{}c@{}}Target ML\\ Trained on\end{tabular}}}} & \multirow{2}{*}{\textit{\textbf{\begin{tabular}[c]{@{}c@{}}Attack\\ Models\end{tabular}}}} & \multicolumn{3}{c}{\textit{Gender}} & \multicolumn{3}{c}{\textit{Age}} & \multicolumn{3}{c}{\textit{Income}} \\ \cline{4-12} 
 &  &  & \textbf{F1-macro} & \textbf{Precision} & \textbf{Recall} & \textbf{F1-macro} & \textbf{Precision} & \textbf{Recall} & \textbf{F1-macro} & \textbf{Precision} & \textbf{Recall} \\ \hline
\multirow{4}{*}{\textit{\textbf{\begin{tabular}[c]{@{}c@{}}Inclusive \\ individual \\ (2013)\end{tabular}}}} & \multirow{2}{*}{Original} & \textit{\begin{tabular}[c]{@{}c@{}}Marginals\\ Only\end{tabular}} & \begin{tabular}[c]{@{}c@{}}0.4976\\$\pm$ 0.0094\end{tabular} & \begin{tabular}[c]{@{}c@{}}0.4977\\$\pm$ 0.0094\end{tabular} & \begin{tabular}[c]{@{}c@{}}0.4977\\$\pm$ 0.0094\end{tabular} & \begin{tabular}[c]{@{}c@{}}0.1237\\$\pm$ 0.0068\end{tabular} & \begin{tabular}[c]{@{}c@{}}0.1238\\$\pm$ 0.0068\end{tabular} & \begin{tabular}[c]{@{}c@{}}0.1238\\$\pm$ 0.0068\end{tabular} & \begin{tabular}[c]{@{}c@{}}0.1982\\$\pm$ 0.0053\end{tabular} & \begin{tabular}[c]{@{}c@{}}0.1982\\$\pm$ 0.0052\end{tabular} & \begin{tabular}[c]{@{}c@{}}0.1983\\$\pm$ 0.0053\end{tabular} \\ \cline{3-12} 
 &  & \textit{\begin{tabular}[c]{@{}c@{}}LOMIA +\\ Marginals\end{tabular}} & \begin{tabular}[c]{@{}c@{}}0.5155\\$\pm$ 0.0080\end{tabular} & \begin{tabular}[c]{@{}c@{}}0.5157\\$\pm$ 0.0081\end{tabular} & \begin{tabular}[c]{@{}c@{}}0.5157\\$\pm$ 0.0080\end{tabular} & \begin{tabular}[c]{@{}c@{}}0.1335\\$\pm$ 0.0053\end{tabular} & \begin{tabular}[c]{@{}c@{}}0.1336\\$\pm$ 0.0052\end{tabular} & \begin{tabular}[c]{@{}c@{}}0.1337\\$\pm$ 0.0053\end{tabular} & \begin{tabular}[c]{@{}c@{}}0.2105\\$\pm$ 0.0072\end{tabular} & \begin{tabular}[c]{@{}c@{}}0.2105\\$\pm$ 0.0071\end{tabular} & \begin{tabular}[c]{@{}c@{}}0.2106\\$\pm$ 0.0072\end{tabular} \\ \cline{2-12} 
 & \multirow{2}{*}{\begin{tabular}[c]{@{}c@{}}Synthetic\\ (CART\\ model)\end{tabular}} & \textit{\begin{tabular}[c]{@{}c@{}}Marginals\\ Only\end{tabular}} & \begin{tabular}[c]{@{}c@{}}0.5035\\$\pm$ 0.0072\end{tabular} & \begin{tabular}[c]{@{}c@{}}0.5036\\$\pm$ 0.0072\end{tabular} & \begin{tabular}[c]{@{}c@{}}0.5036\\$\pm$ 0.0072\end{tabular} & \begin{tabular}[c]{@{}c@{}}0.1227\\$\pm$ 0.0054\end{tabular} & \begin{tabular}[c]{@{}c@{}}0.1228\\$\pm$ 0.0055\end{tabular} & \begin{tabular}[c]{@{}c@{}}0.1227\\$\pm$ 0.0053\end{tabular} & \begin{tabular}[c]{@{}c@{}}0.2020\\$\pm$ 0.0081\end{tabular} & \begin{tabular}[c]{@{}c@{}}0.2021\\$\pm$ 0.0081\end{tabular} & \begin{tabular}[c]{@{}c@{}}0.2020\\$\pm$ 0.0081\end{tabular} \\ \cline{3-12} 
 &  & \textit{\begin{tabular}[c]{@{}c@{}}LOMIA +\\ Marginals\end{tabular}} & \begin{tabular}[c]{@{}c@{}}0.4979\\$\pm$ 0.0086\end{tabular} & \begin{tabular}[c]{@{}c@{}}0.4980\\$\pm$ 0.0087\end{tabular} & \begin{tabular}[c]{@{}c@{}}0.4980\\$\pm$ 0.0087\end{tabular} & \begin{tabular}[c]{@{}c@{}}0.1259\\$\pm$ 0.0057\end{tabular} & \begin{tabular}[c]{@{}c@{}}0.1261\\$\pm$ 0.0057\end{tabular} & \begin{tabular}[c]{@{}c@{}}0.1261\\$\pm$ 0.0057\end{tabular} & \begin{tabular}[c]{@{}c@{}}0.1994\\$\pm$ 0.0082\end{tabular} & \begin{tabular}[c]{@{}c@{}}0.1995\\$\pm$ 0.0082\end{tabular} & \begin{tabular}[c]{@{}c@{}}0.1995\\$\pm$ 0.0082\end{tabular} \\ \hline
\end{tabular}%
}
\end{table}

For completeness, we present the results of LOMIA + Marginals and Marginals Only attacks when adversary resources are Inclusive individuals (2015) (Table~\ref{tab:InferenceResIncl2015}) and Exclusive individuals (2015) (Table~\ref{tab:InferenceResExclu2015}).
Here, on the original data, the LOMIA + Marginals attack is not always more successful than the Marginals Only attack.
However, we do see the trend that attacks are generally slightly less successful when the model is trained on synthetic data. 
We note the risk of leakage posed by the released marginals, and that, moving forward, the danger of releasing marginals must be studied alongside the danger of releasing the ML model itself.

\begin{table}[!htb]
\centering
\caption{Case of ``Inclusive individuals (2015)'' as adversary resources: Evaluation results for model inversion attribute inference attacks using Marginals Only and LOMIA + Marginals attacks. Standard deviations, indicated by $\pm$, represent variability across ten experiment runs.}
\label{tab:InferenceResIncl2015}
\resizebox{\columnwidth}{!}{%
\begin{tabular}{cccccccccccc}
\hline
\multirow{2}{*}{\textit{\textbf{\begin{tabular}[c]{@{}c@{}}Attacker \\ Resources\end{tabular}}}} & \multirow{2}{*}{\textit{\textbf{\begin{tabular}[c]{@{}c@{}}Target ML\\ trained on\end{tabular}}}} & \multirow{2}{*}{\textit{\textbf{\begin{tabular}[c]{@{}c@{}}Attack\\ Models\end{tabular}}}} & \multicolumn{3}{c}{\textit{Gender}} & \multicolumn{3}{c}{\textit{Age}} & \multicolumn{3}{c}{\textit{Income}} \\ \cline{4-12} 
 &  &  & \textbf{F1-macro} & \textbf{Precision} & \textbf{Recall} & \textbf{F1-macro} & \textbf{Precision} & \textbf{Recall} & \textbf{F1-macro} & \textbf{Precision} & \textbf{Recall} \\ \hline
\multirow{4}{*}{\textit{\textbf{\begin{tabular}[c]{@{}c@{}}Inclusive \\ individual \\ (2015)\end{tabular}}}} & \multirow{2}{*}{Original} & \textit{\begin{tabular}[c]{@{}c@{}}Marginals\\ Only\end{tabular}} & \begin{tabular}[c]{@{}c@{}}0.5029\\$\pm$ 0.0077\end{tabular} & \begin{tabular}[c]{@{}c@{}}0.5029\\$\pm$ 0.0076\end{tabular} & \begin{tabular}[c]{@{}c@{}}0.5029\\$\pm$ 0.0076\end{tabular} & \begin{tabular}[c]{@{}c@{}}0.1239\\$\pm$ 0.0065\end{tabular} & \begin{tabular}[c]{@{}c@{}}0.1244\\$\pm$ 0.0064\end{tabular} & \begin{tabular}[c]{@{}c@{}}0.1241\\$\pm$ 0.0067\end{tabular} & \begin{tabular}[c]{@{}c@{}}0.1988\\$\pm$ 0.0092\end{tabular} & \begin{tabular}[c]{@{}c@{}}0.1991\\$\pm$ 0.0090\end{tabular} & \begin{tabular}[c]{@{}c@{}}0.1991\\$\pm$ 0.0091\end{tabular} \\ \cline{3-12} 
 &  & \textit{\begin{tabular}[c]{@{}c@{}}LOMIA +\\ Marginals\end{tabular}} & \begin{tabular}[c]{@{}c@{}}0.5034\\$\pm$ 0.0124\end{tabular} & \begin{tabular}[c]{@{}c@{}}0.5035\\$\pm$ 0.0123\end{tabular} & \begin{tabular}[c]{@{}c@{}}0.5035\\$\pm$ 0.0123\end{tabular} & \begin{tabular}[c]{@{}c@{}}0.1287\\$\pm$ 0.0061\end{tabular} & \begin{tabular}[c]{@{}c@{}}0.1291\\$\pm$ 0.0062\end{tabular} & \begin{tabular}[c]{@{}c@{}}0.1291\\$\pm$ 0.0061\end{tabular} & \begin{tabular}[c]{@{}c@{}}0.1980\\$\pm$ 0.0070\end{tabular} & \begin{tabular}[c]{@{}c@{}}0.1983\\$\pm$ 0.0070\end{tabular} & \begin{tabular}[c]{@{}c@{}}0.1984\\$\pm$ 0.0070\end{tabular} \\ \cline{2-12} 
 & \multirow{2}{*}{\begin{tabular}[c]{@{}c@{}}Synthetic\\ (CART\\ model)\end{tabular}} & \textit{\begin{tabular}[c]{@{}c@{}}Marginals\\ Only\end{tabular}} & \begin{tabular}[c]{@{}c@{}}0.4937\\$\pm$ 0.0083\end{tabular} & \begin{tabular}[c]{@{}c@{}}0.4938\\$\pm$ 0.0082\end{tabular} & \begin{tabular}[c]{@{}c@{}}0.4938\\$\pm$ 0.0082\end{tabular} & \begin{tabular}[c]{@{}c@{}}0.1222\\$\pm$ 0.0055\end{tabular} & \begin{tabular}[c]{@{}c@{}}0.1225\\$\pm$ 0.0057\end{tabular} & \begin{tabular}[c]{@{}c@{}}0.1225\\$\pm$ 0.0053\end{tabular} & \begin{tabular}[c]{@{}c@{}}0.2031\\$\pm$ 0.0066\end{tabular} & \begin{tabular}[c]{@{}c@{}}0.2033\\$\pm$ 0.0066\end{tabular} & \begin{tabular}[c]{@{}c@{}}0.2035\\$\pm$ 0.0067\end{tabular} \\ \cline{3-12} 
 &  & \textit{\begin{tabular}[c]{@{}c@{}}LOMIA +\\ Marginals\end{tabular}} & \begin{tabular}[c]{@{}c@{}}0.5001\\$\pm$ 0.0086\end{tabular} & \begin{tabular}[c]{@{}c@{}}0.5003\\$\pm$ 0.0085\end{tabular} & \begin{tabular}[c]{@{}c@{}}0.5003\\$\pm$ 0.0085\end{tabular} & \begin{tabular}[c]{@{}c@{}}0.1278\\$\pm$ 0.0028\end{tabular} & \begin{tabular}[c]{@{}c@{}}0.1282\\$\pm$ 0.0029\end{tabular} & \begin{tabular}[c]{@{}c@{}}0.1281\\$\pm$ 0.0028\end{tabular} & \begin{tabular}[c]{@{}c@{}}0.1969\\$\pm$ 0.0101\end{tabular} & \begin{tabular}[c]{@{}c@{}}0.1972\\$\pm$ 0.0102\end{tabular} & \begin{tabular}[c]{@{}c@{}}0.1972\\$\pm$ 0.0100\end{tabular} \\ \hline
\end{tabular}%
}
\end{table}

\begin{table}[!htb]
\centering
\caption{Case of ``Exclusive individuals (2015)'' as adversary resources: Evaluation results for model inversion attribute inference attacks using Marginals Only and LOMIA + Marginals attacks. Standard deviations, indicated by $\pm$, represent variability across ten experiment runs.}
\label{tab:InferenceResExclu2015}
\resizebox{\columnwidth}{!}{%
\begin{tabular}{cccccccccccc}
\hline
\multirow{2}{*}{\textit{\textbf{\begin{tabular}[c]{@{}c@{}}Attacker \\ resources\end{tabular}}}} & \multirow{2}{*}{\textit{\textbf{\begin{tabular}[c]{@{}c@{}}Target ML\\ trained on\end{tabular}}}} & \multirow{2}{*}{\textit{\textbf{\begin{tabular}[c]{@{}c@{}}Attack\\ Models\end{tabular}}}} & \multicolumn{3}{c}{\textit{Gender}} & \multicolumn{3}{c}{\textit{Age}} & \multicolumn{3}{c}{\textit{Income}} \\ \cline{4-12} 
 &  &  & \textbf{F1-macro} & \textbf{Precision} & \textbf{Recall} & \textbf{F1-macro} & \textbf{Precision} & \textbf{Recall} & \textbf{F1-macro} & \textbf{Precision} & \textbf{Recall} \\ \hline
\multirow{4}{*}{\textit{\textbf{\begin{tabular}[c]{@{}c@{}}Exclusive\\ individual \\ (2015)\end{tabular}}}} & \multirow{2}{*}{\textit{Original}} & \textit{\begin{tabular}[c]{@{}c@{}}Marginals\\ Only\end{tabular}} & \begin{tabular}[c]{@{}c@{}}0.5002\\$\pm$ 0.0125\end{tabular} & \begin{tabular}[c]{@{}c@{}}0.5012\\$\pm$ 0.0126\end{tabular} & \begin{tabular}[c]{@{}c@{}}0.5012\\$\pm$ 0.0127\end{tabular} & \begin{tabular}[c]{@{}c@{}}0.0880\\$\pm$ 0.0031\end{tabular} & \begin{tabular}[c]{@{}c@{}}0.1275\\$\pm$ 0.0037\end{tabular} & \begin{tabular}[c]{@{}c@{}}0.1323\\$\pm$ 0.0185\end{tabular} & \begin{tabular}[c]{@{}c@{}}0.1504\\$\pm$ 0.0064\end{tabular} & \begin{tabular}[c]{@{}c@{}}0.2001\\$\pm$ 0.0059\end{tabular} & \begin{tabular}[c]{@{}c@{}}0.2027\\$\pm$ 0.0115\end{tabular} \\ \cline{3-12} 
 &  & \textit{\begin{tabular}[c]{@{}c@{}}LOMIA +\\ Marginals\end{tabular}} & \begin{tabular}[c]{@{}c@{}}0.5007\\$\pm$ 0.0065\end{tabular} & \begin{tabular}[c]{@{}c@{}}0.5014\\$\pm$ 0.0066\end{tabular} & \begin{tabular}[c]{@{}c@{}}0.5014\\$\pm$ 0.0066\end{tabular} & \begin{tabular}[c]{@{}c@{}}0.0854\\$\pm$ 0.0055\end{tabular} & \begin{tabular}[c]{@{}c@{}}0.1234\\$\pm$ 0.0050\end{tabular} & \begin{tabular}[c]{@{}c@{}}0.1269\\$\pm$ 0.0262\end{tabular} & \begin{tabular}[c]{@{}c@{}}0.1506\\$\pm$ 0.0065\end{tabular} & \begin{tabular}[c]{@{}c@{}}0.2005\\$\pm$ 0.0052\end{tabular} & \begin{tabular}[c]{@{}c@{}}0.2008\\$\pm$ 0.0123\end{tabular} \\ \cline{2-12} 
 & \multirow{2}{*}{\begin{tabular}[c]{@{}c@{}}Synthetic\\ (CART\\ model)\end{tabular}} & \textit{\begin{tabular}[c]{@{}c@{}}Marginals\\ Only\end{tabular}} & \begin{tabular}[c]{@{}c@{}}0.4966\\$\pm$ 0.0078\end{tabular} & \begin{tabular}[c]{@{}c@{}}0.4979\\$\pm$ 0.0076\end{tabular} & \begin{tabular}[c]{@{}c@{}}0.4979\\$\pm$ 0.0076\end{tabular} & \begin{tabular}[c]{@{}c@{}}0.0839\\$\pm$ 0.0059\end{tabular} & \begin{tabular}[c]{@{}c@{}}0.1233\\$\pm$ 0.0053\end{tabular} & \begin{tabular}[c]{@{}c@{}}0.1264\\$\pm$ 0.0213\end{tabular} & \begin{tabular}[c]{@{}c@{}}0.1447\\$\pm$ 0.0058\end{tabular} & \begin{tabular}[c]{@{}c@{}}0.1980\\$\pm$ 0.0058\end{tabular} & \begin{tabular}[c]{@{}c@{}}0.1952\\$\pm$ 0.0097\end{tabular} \\ \cline{3-12} 
 &  & \textit{\begin{tabular}[c]{@{}c@{}}LOMIA +\\ Marginals\end{tabular}} & \begin{tabular}[c]{@{}c@{}}0.4975\\$\pm$ 0.0078\end{tabular} & \begin{tabular}[c]{@{}c@{}}0.4989\\$\pm$ 0.0078\end{tabular} & \begin{tabular}[c]{@{}c@{}}0.4989\\$\pm$ 0.0079\end{tabular} & \begin{tabular}[c]{@{}c@{}}0.0852\\$\pm$ 0.0030\end{tabular} & \begin{tabular}[c]{@{}c@{}}0.1252\\$\pm$ 0.0025\end{tabular} & \begin{tabular}[c]{@{}c@{}}0.1242\\$\pm$ 0.0146\end{tabular} & \begin{tabular}[c]{@{}c@{}}0.1461\\$\pm$ 0.0075\end{tabular} & \begin{tabular}[c]{@{}c@{}}0.1985\\$\pm$ 0.0068\end{tabular} & \begin{tabular}[c]{@{}c@{}}0.1992\\$\pm$ 0.0140\end{tabular} \\ \hline
\end{tabular}%
}
\end{table}

\section{Conclusion and Future Work}
In this paper, we have investigated an attack on a machine learning model trained to predict an individual's propensity to move i.e., whether they will relocate in the next two years.
We have studied the risk for Inclusive individuals, who are in the training data of the model, as well as for ``unseen'' Exclusive individuals.

To explore the ability of synthetic data to replace original data and protect against model inversion attribute inference attacks, we created fully synthetic data using a CART model.
The ML model trained on the synthetic data maintained prediction performance and was found to leak in the same way or slightly less than the original classifier.
This result is interesting as it shows that training a model on synthetic data will not exacerbate leaks, and may actually have the potential to reduce attribute disclosure risk.
Also, our findings are interesting because until now the SDC community working with synthetic data has mainly focused on measuring the risk of identity disclosure rather than attribute disclosure~\cite{Taub2018Differential}.
In the identity disclosure literature, synthetic data has been shown to provide protection~\cite{domingo2008survey,templ2017statistical}. 
However, releasing a model trained on synthetic data remains an open domain for research.
Our work has highlighted the importance of considering the released marginals and not just the model.

Broadening the scope of the threat model is an essential avenue for future research (Section~\ref{sec:threat}). Exploring additional attack scenarios, such as scenarios involving an attacker with access to confidence scores or confusion matrix from the target machine learning model or scenarios where the attacker lacks access to certain attributes within the data, would contribute to a better understanding of potential vulnerabilities associated with making a trained model publicly available.
In terms of evaluation, future work should consider alternative metrics~\cite{Hittmeir2020Baseline} from both statistical disclosure control (SDC) and machine learning perspectives to evaluate and quantify the success of model inversion attribute inference attacks for a given target individual.
Also, it would be interesting to explore different synthesis approaches ranging from ML and generative models. 
If the inference attack is still possible, then, a second protection using privacy-preserving techniques on sensitive attributes during synthesis, e.g., data perturbation or masking sensitive attributes, might provide extra protection and reduce the risk of attribute disclosure.
Furthermore, the choice of sensitive attributes is important given its impact on the output of model inversion attribute inference attacks.
This consideration extends to understanding the nature of the relationship between sensitive attributes and the target attribute within the machine learning model.

\bibliographystyle{splncs04}
\bibliography{paper.bib}

\end{document}